\pdfoutput=1

\documentclass[11pt]{article}

\usepackage[final]{acl}
\usepackage{booktabs}
\usepackage{xcolor}
\usepackage{times}
\usepackage{latexsym}
\usepackage{placeins}
\usepackage{float}
\usepackage{amsmath}
\usepackage{enumitem}
\usepackage{pgf-pie}
\usepackage{tcolorbox}
\usepackage[T2A,T1]{fontenc} 
\usepackage[utf8]{inputenc}
\usepackage[russian,english]{babel} 
\usepackage{pgfplots}
\usepackage{natbib}

\definecolor{green1}{HTML}{5ACEA0}
\definecolor{orange1}{HTML}{FEDFBE}
\definecolor{blue1}{HTML}{B1E8F5}
\definecolor{purple1}{HTML}{D8C8E1}
\definecolor{yellow1}{HTML}{F5F9C3}

\usepackage[utf8]{inputenc}

\usepackage{microtype}

\usepackage{colortbl}

\usepackage{inconsolata}

\usepackage{graphicx}

\usepackage{pifont}

\title{Towards Open Foundation Language Model and Corpus for Macedonian: A Low-Resource Language}

\author{
 \textbf{Stefan Krsteski\textsuperscript{1}},
 \textbf{Matea Tashkovska$^{*}$\textsuperscript{1}},
 \textbf{Borjan Sazdov$^{*}$\textsuperscript{2,3}},
 \textbf{Hristijan Gjoreski\textsuperscript{2,3}},
\\
 \textbf{Branislav Gerazov\textsuperscript{2}},
\\
 \textsuperscript{1}École Polytechnique Fédérale de Lausanne (EPFL), Switzerland, 
 \\
 \textsuperscript{2} Faculty of Electrical Engineering and Information Technologies, UKIM, North Macedonia,
 \\
 \textsuperscript{3}Emteq Ltd., Brighton, United Kingdom,
\\
  \small{$^{*}$Equal contribution} \\
 \small{
   \textbf{Correspondence:} \href{mailto:email@domain}{stefan.krsteski@epfl.ch}
 }
}

\begin{document}
\maketitle
\begin{abstract}
The increase in technological adoption worldwide comes with demands for novel tools to be used by the general population. Large Language Models (LLMs) provide a great opportunity in this respect, but their capabilities remain limited for low-resource languages, restricting applications in countries where such languages are spoken. We create several resources to facilitate the adoption of LLMs and to support research advancements for Macedonian. We collect the largest Macedonian corpus to date, consisting of 40GB of textual data and totaling 3.5B words. To support conversational applications, we collect a 106k-instance instruction dataset, carefully built to be culturally grounded. For evaluation, we construct a Macedonian evaluation suite covering seven benchmarks. Finally, we train \textit{domestic-yak}, a state-of-the-art 8B-parameter model, on our curated datasets and evaluate it against eight baseline models using the newly constructed benchmark suite. Our model outperforms all existing models in the 8B parameter range across all benchmarks, and achieves performance comparable to models up to 10× larger. Furthermore, a qualitative analysis with native speakers reveals that our model is preferred over larger counterparts, receiving higher ratings for grammatical correctness and cultural appropriateness. All datasets, code, and model weights are openly released, setting a foundation for advancing LLMs in similarly underrepresented languages. These resources are publicly available at \href{https://github.com/LVSTCK}{\texttt{github.com/LVSTCK}} for source code, and at \href{https://huggingface.co/LVSTCK}{\texttt{huggingface.co/LVSTCK}} for pretrained model weights and data.

\end{abstract}

\section{Introduction}
As Large Language Models (LLMs) continue to transform modern natural language processing (NLP), the benefits of these advances remain disproportionately concentrated among high-resource languages~\cite{joshi2020state}. With over 7,000 languages spoken globally, most remain severely underrepresented in the training data that powers these models, limiting access to AI for billions of people worldwide~\cite{blasi2021systematic}.

Despite the development of multilingual variants aimed at addressing this disparity, significant challenges remain for low-resource languages. These models often lack the depth of understanding necessary for high-quality performance across all languages they claim to support. This issue is particularly pronounced in languages with smaller speaker populations, such as Macedonian, which belongs to the Eastern South Slavic branch and is spoken by over 1.1 million native speakers~\cite{makcensus2021}. The fundamental relationship between data quantity and model performance means that languages with limited representation in training corpora inevitably experience degraded results~\cite{kaplan2020scaling}. In addition, the absence of standardized evaluation benchmarks makes progress difficult to measure.

In this work, we present a thorough approach to advancing Macedonian NLP through the development of several language-specific resources:

\begin{enumerate}
\setlength\itemsep{0.1em}
\item The largest Macedonian corpus to date is collected, consisting of over 3.5 billion words aggregated from four existing and eight newly collected sources.
\item A novel Macedonian instruction-tuning dataset is constructed, featuring multi-turn dialogue, synthetic, commonsense, and logical reasoning examples, refined through human feedback and LLM-assisted filtering.
\item We introduce \textbf{domestic-yak}, an 8B-parameter foundation language model for Macedonian, offering both pretrained and instruction-tuned variants. It outperforms existing models in its class and achieves performance comparable to models 10$\times$ larger.
\item We create an evaluation benchmark designed to assess model performance in Macedonian across multiple tasks, including commonsense reasoning and reading comprehension.
\end{enumerate}

Our results demonstrate that targeted, language-specific development can significantly help in increasing performance. By open-sourcing all data, code, and model weights, we hope to contribute both immediate value to the Macedonian-speaking community and a reproducible blueprint for similar efforts in other languages. 

The paper is structured as follows: Section 2 reviews related work. Section 3 details our data collection methodology. Section 4 describes the model training setup. Section 5 presents the evaluation framework, covering both quantitative and qualitative setups. Section 6 reports and discusses the results. Section 7 ends the paper with a conclusion.

\section{Related Work}
\paragraph{Corpora for Multilingual Models.}

A key factor contributing to the success of LLMs in English has been the wide availability of high-quality text resources, as performance improvements correlate strongly with both corpus size and quality~\cite{kaplan2020scaling}. Large-scale English corpora such as Common Crawl\footnote{\url{https://commoncrawl.org}}, The Pile~\cite{pile}, and C4~\cite{raffel2020exploring} have provided the scale and variety needed to train increasingly capable models. These datasets offer large volumes of text and cover many domains, styles, and linguistic features, making them effective for pretraining and thereby enabling better knowledge transfer to downstream tasks.

In contrast, many low-resource languages lack such large, comprehensive corpora, creating a significant barrier to the development of competitive language models. However, in recent years, there have been growing efforts to close this gap through creating multilingual datasets that aggregate content across languages and enable training at a global scale.  Similar to the English-centric datasets, multilingual resources such as mC4 \cite{xue2020mt5}, OSCAR~\cite{suarez2020monolingual}, Fineweb2~\cite{fineweb-2} and HPLT-v2~\cite{hplt2}, have been introduced to facilitate large-scale pretraining. These datasets aim to provide a better foundation for building models that generalize across a wider range of languages and cultures. 

Nevertheless, even within these multilingual collections, representation remains uneven. High-resource languages dominate the data distribution, while low-resource Slavic languages like Macedonian are often underrepresented, both in terms of quantity and quality. To address this gap, we introduce an open-source corpus designed to advance research for this underrepresented language.

\paragraph{Language Modeling Approaches.}
The availability of multilingual datasets has enabled a shift from English-centric to multilingual models. For instance, models such as GPT-4o and Llama-3 now claim native support for over 100 languages~\cite{grattafiori2024llama}. Beyond these efforts, researchers have investigated more efficient strategies to extend existing models to low-resource languages. Parameter-efficient approaches, such as MAD-X~\cite{pfeiffer2020mad}, incorporate lightweight language and task adapters to enable zero-shot transfer while training only 3–5\% of the model's parameters. Alternatively, large-scale continual pretraining has been shown to introduce hundreds of new languages simultaneously, yielding strong task generalization. For instance, EMMA-500~\cite{ji2024emma}, trained on 546 languages, achieves significant gains without any task-specific fine-tuning.

Alongside general multilingual models, some approaches focus on groups of closely related languages to exploit shared linguistic structure. The CroSloEngual model~\cite{ulcar-robnik2020finest}, for instance, was pretrained from scratch on Croatian, Slovene, and English, aiming to support multi- and cross-lingual training across these languages. Similarly, YugoGPT~\cite{YugoGPT} is a recent effort that trains the best 7B-parameter LLM for Bosnian, Croatian, and Serbian. Furthermore, the BERTić model~\cite{ljubevsic2021berti}, was trained on Bosnian, Croatian, Montenegrin, and Serbian, which are languages that form the pluricentric Serbo-Croatian language and have overlapping vocabulary and grammar. This strategy allows for efficient use of limited data while still benefiting from multilingual learning as the languages share strong structural and lexical similarities.

However, where sufficient high-quality data exists, monolingual models are also emerging as a better alternative. For instance, for Vietnamese, continued pretraining on top of multilingual backbones followed by instruction tuning led to improvements across 10 tasks over multilingual baselines~\cite{truong2024crossing}. Similar trends are seen in recent monolingual models for Italian~\cite{orlando2024minerva}, Arabic~\cite{koubaa2024arabiangpt}, and Finnish~\cite{luukkonen2023fingpt}.

Macedonian remains underrepresented, with only a single publicly available language model to date\footnote{\url{https://huggingface.co/trajkovnikola/MKLLM-7B-Instruct}}. We address this issue with our work and present a new large-scale language model for Macedonian, providing both pretrained and fine-tuned versions.

\section{Data}
In this section, we present two contributed datasets: a Macedonian corpus and an instruction dataset designed to elicit chat capabilities. We describe their properties and explain how they were collected and prepared.

\subsection{Macedonian Corpus}
To construct our corpus, we combine well-established sources with newly published data that have remained unexploited  in Macedonian NLP research. These new sources include academic publications, educational materials spanning elementary to university levels, and various text-rich documents, typically available as PDFs on the web. The sources used are described in detail below and summarized in Table~\ref{tab:dataset_splits}.

\textbf{FineWeb2}~\cite{fineweb-2} represents one of the most popular web crawled datasets available for the non-English community. Sourced from 99 CommonCrawl snapshots that span from 2013 to 2024, the data underwent deduplication and quality filtering. For our purposes, we use only the Macedonian portion of this dataset. 

\textbf{HPLT-v2}~\cite{hplt2} provides another valuable resource in our corpus. This collection includes 193 languages and was derived from web crawls subjected to similar processing as FineWeb2. Similarly, we isolate only the Macedonian subset. 

\textbf{MaCoCu-mk 2.0}~\cite{macocu} represents another well-known web crawl resource. The Macedonian subset was constructed by crawling the ".mk" Internet top-level domains in 2021. 

\textbf{Document-to-Text.}
Historically, pre-training data for Macedonian language models has been sourced from web crawls, as shown by the preceding collections. To expand beyond these limitations, we contribute new data sources that have remained untapped to date. Several tools have recently emerged to facilitate document-to-text conversion, including \textit{docling}~\cite{docling}, \textit{nv-ingest}~\cite{nv-ingest}, and \textit{mmore}~\cite{sallinen2025mmore}. In our work, we use \textit{mmore} to extract high-quality text from a variety of document sources, particularly focusing on academic publications, educational materials, official government documents and other scanned digital resources. More information on these tools and the full list of processed sources is available in the Appendix \ref{sec:corpus}.

\textbf{Wikipedia.}
As a standard resource in language modeling, we include the "mk" Wikipedia dump with the last update being January 2025. 

\textbf{SETimes Corpus}~\cite{ljubevsic2023macedonian}, 
is a parallel corpus of news articles in the  Balkan languages.  In this work we use the complete Macedonian-English pair (207,777 sentence pairs; 44.6M
tokens) and retain only the Macedonian side.

\textbf{Common Voice}~\cite{ardila2019common} is an open-source, multilingual dataset originally developed to train speech-enabled applications. It provides transcriptions in the form of natural text prompts for speakers. We extract only the Macedonian transcription text, which consists of human-validated sentences. Although not originally intended as a text corpus, it offers an unconventional but high-quality source of conversational language.

\begin{table}[h]
\centering
\resizebox{\linewidth}{!}{%
\begin{tabular}{lrr}
\hline
\textbf{Origin} & \textbf{Words (B)} & \textbf{Percentage} \\
\hline
HPLT-2                   & 1.49   & 42.21\%   \\
FineWeb2                & 1.33   & 37.66\%   \\
MaCoCu-mk 2.0            & 0.49   & 13.92\%   \\
Documents (mmore)        & 0.14   & 4.07\%    \\
Wikipedia                & 0.07   & 1.96\%    \\
SETimes Corpus           & 0.004  & 0.13\%    \\
Common Voice             & 0.002  & 0.05\%    \\
\hline
\textbf{Total}           & \textbf{3.53} & \textbf{100.00\%} \\
\hline
\end{tabular}%
}
\caption{Sources and word distribution for the Macedonian pretraining corpus}
\label{tab:dataset_splits}
\end{table}

The resulting corpus consists of 3.53 billion words. Given the significant overlap between web-based sources (particularly those derived from CommonCrawl) and recent evidence demonstrating that filtering and deduplication significantly improve language model performance~\cite{lee2021deduplicating}, we implement a text filtering pipeline, closely following FineWeb2's methodology~\cite{fineweb-2}. 

As an initial step, we remove Personally Identifiable Information (PII) such as email and IP addresses, and telephone numbers to comply with privacy regulations using \textit{datatrove}~\cite{penedo2024datatrove}. We then apply C4 filtering~\cite{raffel2020exploring} to discard low-quality content, including removing lines with fewer than three words or lines lacking terminal punctuation. 

Furthermore, we implement Gopher filtering~\cite{rae2021scaling}, including rejecting instances where over 90\% of lines begin with bullets or where more than 30\% of lines end with ellipses. We use the FastText language identification model~\cite{joulin2016bag, joulin2016fasttext} to retain only high-confidence Macedonian text (confidence > 0.65). Following this, we perform sentence-level deduplication to remove redundant content. For the newly contributed document-based data, we apply sentence chunking to segment texts into manageable units, each not exceeding 4000 words. 

Finally, we use MinHash-based locality-sensitive hashing~\cite{broder1997resemblance} for document-level deduplication, removing near-duplicate documents across the entire corpus. The multistage filtering pipeline resulted in 1.47 billion words of high-quality text.

\subsection{Instruction Dataset}  
\label{sec:instruction-dataset}
Most existing instruction datasets~\cite{upadhayay2024taco} for Macedonian rely on direct translation from English, which introduces both linguistic artifacts and cultural mismatches~\cite{bizzoni-etal-2020-human}. To overcome these limitations, we use a hybrid construction methodology combining human supervision with model-assisted refinement. Specifically, we post-edit translated instances using GPT-4o-mini~\cite{openai2024gpt4ocard}, by instructing it to grammatically refine the translated sentences, followed by human verification to filter low-quality samples. This process enables us to build a richer, culturally appropriate dataset while minimizing translation noise. Our final dataset integrates several sources, each selected to support specific capabilities, which we describe in details below, with the summary available in Table~\ref{tab:dataset_splits_sft_words}.

\textbf{General Instruction Following.}  
To support broad task coverage, we incorporate \textit{Alpaca}~\cite{alpaca} and \textit{Databricks-Dolly}~\cite{DatabricksBlog2023DollyV2}, two well-known instruction datasets. These primarily include instruction-following examples including tasks such as brainstorming, classification, closed and open question answering, generation, information extraction, and summarization. Since both datasets were produced using earlier models (e.g., GPT-3) and translated automatically, the aforementioned refinement was necessary to address issues in fluency and cultural misalignment.

\textbf{Conversational Abilities.}  
To support multi-turn conversational capabilities, we include \textit{UltraChat 200k}~\cite{ding2023enhancing} and \textit{Capybara}~\cite{daniele2023amplify-instruct}. \textit{UltraChat} focuses on assistant-style dialogues across a wide range of user intents, while \textit{Capybara} focuses on multi-turn reasoning, logic and extrapolation about a wide range of subjects. These sources contribute to the conversational fluency of the final dataset. 

\textbf{Reasoning.}  
To incorporate reasoning capabilities, we translated a subset of the  \textit{Open Platypus}~\cite{platypus2023} dataset, which focuses on improving logical reasoning skills in language models. This dataset mainly consists of mathematical problems that challenge the model's reasoning abilities.

\textbf{Culturally Grounded Content.}  
To address the scarcity of Macedonian-specific content and to ensure cultural relevance beyond what translated datasets could provide, we generate \textit{synthetic data}. Using GPT-4o-mini with in-context learning, we create 3,400 culturally relevant input-output pairs across domains such as geography, history, education, science, religion, and governance. These examples are then post-processed and manually reviewed to ensure higher quality.

\begin{table}[h]
    \centering
    \begin{tabular}{lrr}
        \hline
        \textbf{Origin} & \textbf{Words (M)} & \textbf{Percentage} \\
        \hline
        Alpaca$^\dagger$              & 13.01 & 16.95\% \\
        Ultrachat           & 34.14 & 44.48\% \\
        Capybara            & 22.63 & 29.48\% \\
        Databricks Dolly$^\dagger$    & 3.38  & 4.40\%  \\
        Open Platypus$^\dagger$       & 1.80  & 2.34\%  \\
        Synthetic Data$^\dagger$      & 1.80  & 2.34\%  \\
        \hline
        \textbf{Total}      & \textbf{76.76} & \textbf{100.00\%} \\
        \hline
    \end{tabular}
    \caption{Source distribution of the Macedonian instruction-tuning dataset. 
    Datasets marked with $^\dagger$ were refined through model-assisted post-editing and human verification to improve fluency and cultural relevance.}
    \label{tab:dataset_splits_sft_words}
\end{table}

The final instruction dataset contains 106,993 samples and approximately 77 million words, covering tasks such as question answering, chat conversations, mathematical reasoning, essay writing and code generation. Table~\ref{tab:dataset_splits_sft_words} summarizes the dataset composition, while Appendix~\ref{app:instruct_data} (Figure~\ref{fig:instruction_topic_pie}) illustrates the topic distribution.

\section{Language Model Training}
Our training procedure follows a two-stage approach: continued pretraining on raw text (the corpus), followed by supervised fine-tuning (SFT) on instruction data.

\subsection{Continued Pretraining}
In the pre-training stage, the model is optimized to predict the next token in a sequence using the standard autoregressive objective. Given a token sequence $\{x_1, \dots, x_T\}$, the training objective is to maximize the log-likelihood:

\begin{equation}
\mathcal{L} = \sum_{t=1}^{T} \log P(x_t \mid x_{<t})
\end{equation}

where $T$ denotes the sequence length, $x_t$ is the token at position $t$, and $x_{<t}$ represents the preceding tokens.

Rather than training from scratch on our corpus, we continue pre-training from the publicly available \textit{Llama3.1 8B Instruct} model weights. This approach exploits the knowledge learned during the models' original multilingual training, which is especially useful for low-resource settings where data scarcity is a major bottleneck~\cite{ji2024emma}. We retain the original tokenizer to avoid the complexity of re-tokenization. Training spans for one epoch over the full corpus using four H100 GPUs (80 GB each 320 GB total). We use a maximum sequence length of 8{,}192 tokens, a cosine annealing scheduler (peak learning rate $2\times10^{-5}$), and the AdamW optimizer. To optimize memory usage, we set a per-device batch size of 1 and use gradient accumulation over 8 steps.

\subsection{Supervised Fine-Tuning}
Full fine-tuning is performed on top of our pretrained model using the instruction dataset. To make use of higher quality data, we sample with a 2:1 sampling ratio favoring human-supervised and synthetic examples over translated ones. Based on an analysis of the instruction lengths, we set the maximum sequence length to 4{,}096 tokens, covering over 95\% of the dataset without truncation (see Appendix~\ref{app:instruct_data}, Figure~\ref{fig:cdf}). We optimize the standard cross-entropy loss over the instruction data, i.e. negative log-likelihood of the next token given the prefix. Training spans for three epochs using a single H100 GPU (80 GB). We use the AdamW optimizer with a per-device batch size of 2 and gradient accumulation over 8 steps. We double the learning rate to $4 \times 10^{-5}$ and use the same scheduling method as in the pre-training phase.

\section{Evaluation Setup} 

\begin{table*}[t]
\centering
\small
\begin{tabular}{llcccccccc}
\toprule
\textbf{Model} & \textbf{Size} & \textbf{PIQA} & \textbf{OBQA} & \textbf{WinoG} & \textbf{ARC-E} & \textbf{ARC-C} & \textbf{BoolQ} & \textbf{HSwag} & \cellcolor{gray!20}\textbf{Avg.} \\
\midrule
\multicolumn{10}{c}{\textit{Smaller Models}} \\
\midrule
Llama 3.2 & 1B & 0.539 & 0.162 & 0.509 & 0.231 & 0.190 & 0.573 & 0.270 & \cellcolor{gray!20}0.353 \\
Phi-3.5-mini & 3.8B & 0.526 & 0.164 & 0.519 & 0.289 & 0.188 & 0.603 & 0.263 & \cellcolor{gray!20}0.364 \\
\midrule
\multicolumn{10}{c}{\textit{Comparable Sizes (7B--8B)}} \\
\midrule
Qwen2.5 & 7B & 0.560 & 0.216 & 0.535 & 0.391 & 0.253 & 0.779 & 0.339 & \cellcolor{gray!20}0.439 \\
Mistral & 7B & 0.578 & 0.218 & 0.561 & 0.463 & 0.287 & 0.759 & 0.372 & \cellcolor{gray!20}0.462 \\
Llama 3.1 & 8B & 0.587 & 0.252 & 0.568 & 0.445 & 0.282 & 0.764 & 0.374 & \cellcolor{gray!20}0.467 \\
MKLLM$^\dagger$ & 7B & 0.642 & 0.294 & 0.615 & 0.503 & 0.300 & 0.788 & 0.433 & \cellcolor{gray!20}0.510\\
\rowcolor{green1!30}
\textbf{domestic-yak}$^\dagger$ & 8B & \textbf{0.692} & \textbf{0.302} & \textbf{0.627} & 0.547 & 0.336 & 0.787 & 0.448 & \cellcolor{gray!20}0.535 \\
\midrule
\multicolumn{10}{c}{\textit{Larger Models (12B--70B)}} \\
\midrule
Mistral Nemo & 12B & 0.607 & 0.242 & 0.606 & 0.472 & 0.319 & 0.809 & 0.400 & \cellcolor{gray!20}0.493 \\
\rowcolor{blue1!40}
Llama 3.3 & 70B & 0.660 & 0.282 & 0.609 & \textbf{0.581} & \textbf{0.369} & \textbf{0.851} & \textbf{0.466} & \cellcolor{gray!20}0.545 \\
\bottomrule
\end{tabular}
\caption{
Performance comparison across models (all in their instruction-tuned variants), evaluated with accuracy. Benchmarks are sorted by average score (descending) within each model class.
Models with explicit support for Macedonian are marked with $^\dagger$. For the remaining models, we could not confirm language coverage. Despite being over 10$\times$ smaller, our 8B model outperforms Llama 70B on 3 out of 7 benchmarks (PIQA, OBQA, WinoG). Standard deviations were consistent (0.009–0.014) and are omitted for clarity.
}
\label{tab:model-comparison}
\end{table*}

\subsection{Benchmarks} 
Similar to many other low-resource languages, Macedonian lacks a standardized evaluation benchmark, making it difficult to track progress in LLM development. To address this, we construct a Macedonian adaptation of the Language Model Evaluation Harness~\cite{eval-harness}. 

A natural approach would be to translate the original English benchmarks directly into Macedonian. However, as discussed in Section~\ref{sec:instruction-dataset}, translations from English tend to introduce unnatural phrasing, so called "translationese", and cultural biases, which can make the benchmarks unreliable for evaluating models in the target language. To address these issues, we instead leverage an existing high-quality benchmark adaptation available for Serbian~\cite{serbian-llm-eval}. Given the close linguistic and cultural affinities between these two South Slavic languages, we translate the Serbian version into Macedonian, maintaining natural phrasing and improving evaluation fidelity. 

Furthermore, to preserve grammatical correctness during translation, we use a template-based strategy. Translating individual text segments (multiple-choice questions without answer options) often disrupts target language word order. To address this, we translate full sentence templates containing placeholders for answer options, then remove the placeholders post-translation. See Appendix~\ref{sec:translation} for implementation details and examples. 

In total, we translated seven benchmarks, which we use to quantitatively measure the performance of our model using accuracy as the evaluation metric. The benchmarks cover two task categories: commonsense reasoning and reading comprehension.

\textbf{Commonsense Reasoning} benchmarks evaluate an LLM’s ability to apply everyday human-like assumptions that are not explicitly stated. This includes physical world knowledge, causal and temporal reasoning, as well as understanding of social norms and expectations. We report results on six well-known datasets (in their translated versions): \textit{HellaSwag}~\cite{hellaswag}, \textit{WinoGrande}~\cite{winogrande}, \textit{PIQA}~\cite{piqa}, \textit{OpenbookQA}~\cite{OpenBookQA2018}, \textit{ARC-Easy}, and \textit{ARC-Challenge}~\cite{allenai:arc}.

\textbf{Reading Comprehension} benchmarks evaluate the ability of a model to understand a given text passage, specifically its ability to grasp context, coherence and narrative flow. We evaluate performance using the \textit{BoolQ} dataset~\cite{boolq}. 

\subsection{Qualitative Evaluation} 
In addition to quantitative evaluation, we conduct an analysis where we assess the quality of responses through native speaker judgments. We carry out a head-to-head comparison between our \textit{domestic-yak} and the strongest evaluated model Llama 3.1 70B. We design ten original questions (included in Appendix~\ref{appendix:questions}) that reflect everyday reasoning, culturally grounded knowledge, and typical native language use. Specific tasks include understanding common expressions, giving advice, writing informal messages, and answering questions about local institutions. For each question, native speakers evaluate the responses of both models. A total of 35 participants completed the survey, with a mean age of 28$\pm$9 years, including 19 males and 16 females. 
Participants evaluated each pair of responses by selecting the better answer and providing a brief justification. The available options for the justification included better gramatical consistency, more natural phrasing, higher cultural appropriateness, more information, and an open ``Other'' field for free-text input. Moreover, the participants rated both of the answers for fluency and relevance using a Likert scale from 1 to 5~\cite{likert1932technique}. To reduce bias, model outputs were anonymized and randomized across questions, with responses labeled as ``Model A'' and ``Model B''. The goal of this human evaluation is to highlight differences that are not captured by quantitative benchmarks alone.

\section{Results and Discussion}
\subsection{Quantitative Results}

We compare \textit{domestic-yak-instruct} (8B) against eight baselines spanning three size categories: smaller (1B–4B), in-class (7B–8B), and larger (12B–70B). The results are shown in Table~\ref{tab:model-comparison}. Three key takeaways emerge from this comparison.
Firstly, our model achieves the highest performance among all models of comparable size across every evaluated task, outperforming strong baselines such as Mistral, Qwen2.5, and Llama 3.1. We attribute this significant improvement to our targeted training strategies, particularly the use of the largest Macedonian corpus combined with the instruction dataset that enables the model to better capture the linguistic patterns. Secondly, \textit{domestic-yak} outperforms larger counterparts, surpassing Mistral Nemo (12B) on all but one task, and Llama-3.3 (70B) on three of seven benchmarks (\textit{PIQA}, \textit{OpenBookQA}, \textit{WinoGrande}), despite being an order of magnitude smaller. Finally, our model represents a significant improvement compared to the previous best Macedonian model, MKLLM, achieving higher accuracy across six out of seven benchmarks. In summary, \textit{domestic-yak} sets a new state-of-the-art result for the Macedonian language and marks a significant step forward for NLP in this domain, laying the foundation for a full suite of models that will be released in the near future.

\subsection{Ablation Study}
\begin{table*}[t]
\small
\centering
\begin{tabular}{lcccc}
\toprule
\textbf{Task (mk)} 
& \textbf{Llama 3.1} 
& \textbf{domestic-yak-base (+ $\Delta_1$)} 
& \textbf{domestic-yak-instruct (+ $\Delta_2$)} 
& \textbf{total $\Delta$} \\
\midrule
ARC Easy        & 0.45 & 0.52 \textcolor{green1!80!black}{(+0.07)} & 0.55 \textcolor{green1!80!black}{(+0.03)} & \textbf{\textcolor{green1!80!black}{+0.10}} \\
ARC Challenge   & 0.28 & 0.32 \textcolor{green1!80!black}{(+0.04)} & 0.34 \textcolor{green1!80!black}{(+0.02)} & \textbf{\textcolor{green1!80!black}{+0.06}} \\
BoolQ           & 0.76 & 0.77 \textcolor{green1!80!black}{(+0.01)} & 0.79 \textcolor{green1!80!black}{(+0.02)} & \textbf{\textcolor{green1!80!black}{+0.03}} \\
HellaSwag       & 0.37 & 0.43 \textcolor{green1!80!black}{(+0.06)} & 0.45 \textcolor{green1!80!black}{(+0.02)} & \textbf{\textcolor{green1!80!black}{+0.08}} \\
Openbook QA     & 0.25 & 0.29 \textcolor{green1!80!black}{(+0.04)} & 0.30 \textcolor{green1!80!black}{(+0.01)} & \textbf{\textcolor{green1!80!black}{+0.05}} \\
PIQA            & 0.59 & 0.67 \textcolor{green1!80!black}{(+0.08)} & 0.69 \textcolor{green1!80!black}{(+0.02)} & \textbf{\textcolor{green1!80!black}{+0.10}} \\
WinoGrande      & 0.57 & 0.63 \textcolor{green1!80!black}{(+0.06)} & 0.63 \textcolor{gray!50!black}{(+0.00)} & \textbf{\textcolor{green1!80!black}{+0.06}} \\
\midrule
\textbf{Average} & \textbf{0.47} & \textbf{0.52} \textcolor{green1!80!black}{\textbf{(+0.05)}} & \textbf{0.54} \textcolor{green1!80!black}{\textbf{(+0.02)}} & \textbf{\textcolor{green1!80!black}{+0.07}} \\
\bottomrule
\end{tabular}
\caption{
Ablation study on the effects of pre-training and instruction tuning. 
\textbf{Llama 3.1} is the base model. 
\textbf{domestic-yak-base} is a result from continued pretraining using our corpus, and 
\textbf{domestic-yak-instruct} adds instruction tuning. 
We report stepwise improvements inline in \textcolor{green1!80!black}{green}, and total gains are highlighted in \textbf{{\textcolor{green1!80!black}{bold green}}}.
}
\label{tab:ablations}
\end{table*}
A central objective of this work is to demonstrate the effectiveness of our proposed Macedonian corpus and instruction dataset for adapting language models. To break down their impact, we run an ablation study measuring performance gains. Starting from the baseline \textit{Llama-3.1-8B-Instruct} model, we incrementally apply (\emph{i}) continued pre-training on our Macedonian corpus (\textit{domestic-yak-base}), and (\emph{ii}) supervised fine-tuning on the instruction dataset (\textit{domestic-yak-instruct}). Table~\ref{tab:ablations} reports the results, isolating the effects of domain-specific pretraining and instruction tuning.

The pre-training phase provides the majority of gains, increasing the average score from 0.47 to 0.52. Improvements are consistent across all Commonsense Reasoning benchmarks, with \textit{PIQA} (+8), \textit{ARC Easy} (+7), and \textit{HellaSwag} (+6) among the highest. In contrast, the Reading Comprehension benchmark (\textit{BoolQ}) shows only a marginal (+1) improvement. Since many of the tasks with larger improvements primarily test factual recall, this pattern suggests that continued pretraining is very effective at enhancing the model’s factual knowledge. Meanwhile, skills such as contextual reading and coherence tracking appear to be well-covered by the base model, as no significant improvements were seen for that task category. 

Instruction tuning provides an additional +2 points on average. It improves performance on tasks such as \textit{ARC Easy} and \textit{BoolQ}, but has no positive effect on \textit{WinoGrande}, where pronoun-resolution skills (Winogrande's main task) plateau during pretraining. This limited effect is consistent across tasks, which we attribute to the strength of the base model. Since \textit{Llama 3.1 Instruct} is already trained for general-purpose instruction following, additional fine-tuning on task-specific instructions largely acts as light alignment. It helps adapt the model to domain-specific phrasing and task format, but contributes little in terms of new capability.

\subsection{Qualitative Analysis}
We collected human evaluation data comparing responses from \textit{domestic-yak-instruct} and \textit{Llama 3.1 70B Instruct} across ten unique prompts. The analysis includes model preference counts and Likert scale ratings for fluency and relevance. Overall, \textit{domestic-yak-instruct} was preferred in 64.2\% of the pairwise comparisons, while \textit{Llama 3.1 70B Instruct} was preferred in 27.1\%. In 8.7\% of cases, participants rated the two responses equally. 

\begin{figure}[t!]
\centering
\includegraphics[width=0.94\linewidth]{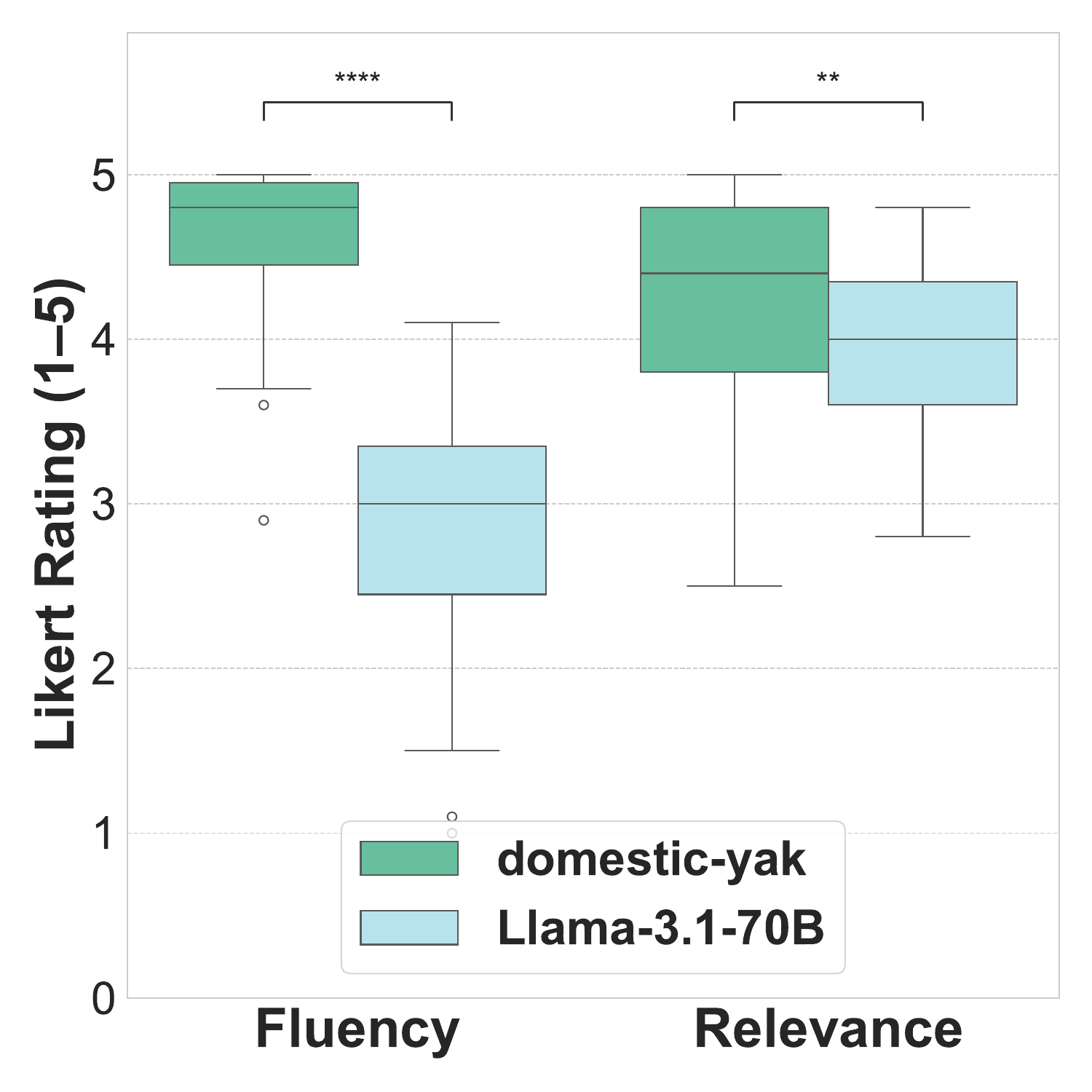}
\caption{Average fluency and relevance Likert ratings per model. \textit{domestic-yak-instruct} outperforms \textit{Llama 3.1 70B Instruct} in both dimensions (Wilcoxon signed-rank test, Bonferroni corrected, $p$\textsubscript{fluency}=1.83$\times$10\textsuperscript{-11}, $p$\textsubscript{relevance}=1.192$\times$10\textsuperscript{-3}). Statistical significance annotations: * if $p \in [0.05, 10^{-2})$; ** if $p \in [10^{-2}, 10^{-3})$; *** if $p \in [10^{-3}, 10^{-4})$; and **** if $p \leq 10^{-4}$.}
\label{fig:likert}
\end{figure}

Participants most cited better grammatical consistency (81.6\%), more natural phrasing (60\%), and higher cultural appropriateness (37\%) as reasons for preferring our model. In Likert ratings, our model achieved average scores of 4.6 for fluency and 4.26 for relevance, compared to 2.8 and 3.9 for \textit{Llama 3.1 70B}, respectively.

To formally test these differences, we grouped the results by participant. For each participant, we computed the number of times each model was preferred and the average Likert ratings for fluency and relevance. A Shapiro--Wilk test indicated that the distributions were not normal, so we applied the Wilcoxon signed-rank test for all comparisons, which tests if the median difference between pairs is zero. Bonferroni correction was used to adjust for multiple testing. Regarding model preference, the results demonstrate a statistically significant difference ($p$ = 8.56$\times$10\textsuperscript{-5}), with participants favoring \textit{domestic-yak-instruct} over \textit{Llama 3.1 70B Instruct}. Similarly, as shown in Figure~\ref{fig:likert}, our model significantly outperformed the baseline in both fluency ($p$ = 9.17$\times$10\textsuperscript{-11}) and relevance ($p$ = 7.26$\times$10\textsuperscript{-3})

Although Llama 70B achieved higher scores on several quantitative benchmarks (Table ~\ref{tab:model-comparison}), our model was highly preferred by native speakers during qualitative evaluation. This demonstrates that benchmark scores do not fully capture the whole story, i.e. real-world, language-specific model quality. By continuing pretraining on high-quality data and applying instruction tuning across a broad range of tasks, including general instruction following, culturally grounded content, reasoning and conversational skills, our model learned the linguistic and cultural characteristics of the Macedonian language crucial for native speakers. The qualitative results confirm that our model surpasses a model nearly ten times larger in fluency, relevance, and overall preference among native speakers, proving that careful adaptation can rival scale (see Appendix~\ref{appendix:questions} for example responses).

\section{Conclusion}

In this work, we bridge the gap in Macedonian NLP by introducing a suite of language-specific resources and demonstrating the effectiveness of focused monolingual adaptation in low-data settings. We release the largest Macedonian corpus (3.5B+ words), a cleaned version of the said dataset (1.5B+ words), a conversational instruction-tuning dataset, and a standardized evaluation benchmark spanning commonsense reasoning, factual knowledge, and reading comprehension. Using these resources, we train and release \textit{domestic-yak}, an 8B-parameter model that outperforms existing baselines and matches or surpasses multilingual models up to ten times larger across tasks.

Ablations highlight the importance of continued monolingual pretraining, which resulted in greater gains than instruction tuning alone, emphasizing the value of high-quality, language-specific data. Human evaluations further strengthen our findings: native speakers consistently preferred \textit{domestic-yak-instruct} over the \textit{Llama 3.1 70B Instruct}, rating it significantly higher for fluency, grammatical accuracy, and cultural relevance.

Our results prove that targeted resource development and monolingual adaptation enable smaller models to outperform larger multilingual systems in real-world applications. All datasets, benchmarks, and model weights are publicly released to accelerate Macedonian NLP research and applications. Future work will expand the benchmark to include broader task coverage and address the current 4k context-length limitation to support applications requiring larger windows. We also plan to incorporate additional datasets, such as COPA-MK~\cite{11356/1687}, a Macedonian translation of the Choice of Plausible Alternatives (COPA) benchmark~\cite{ponti2020xcopa}, as well as resources from the OPUS collection~\cite{TIEDEMANN12.463} to further improve model robustness and evaluation depth. We hope this work offers a blueprint for revitalizing other low-resource languages through targeted efforts, free from the constraint of scale.

\section*{Limitations}
We identify three main limitations in our work. 
First, while the model performs well on general-purpose tasks, it has not been evaluated nor adapted for niche domains such as law, medicine, or finance. Performance in these areas is likely to be limited due to the lack of domain-specific data. Accordingly, we position this release as a general-purpose foundation and encourage the community to pursue fine-tuning and evaluation in specialized domains.

Furthermore, the model uses a maximum context window of 8,192 tokens during pretraining and 4,096 tokens during instruction tuning. This limits its ability to handle tasks that require longer context, such as multi-document summarization or long-form QA. We believe that addressing this limitation should be the key focus of future work, both in data collection and model training processes. 

Finally, we note as a minor limitation the lack of Macedonian benchmarks, which required us to rely on translated datasets. This introduces variance that can negatively affect the accuracy of Macedonian-specific quantitative evaluation, even though we took steps to reduce it. Nevertheless, comparison is made against the same datasets, so this does not significantly reduce the confidence in the presented results.

\newpage
\onecolumn

\appendix
\section{Appendix}
\label{sec:appendix}

\subsection{Document-to-Text}
\label{sec:corpus}

We mention in our main text that a significant portion of our corpus was collected using document-to-text tools. Recently, such tools are well established in the community and enable text extraction from diverse file formats (PDF, DOCX, PPTX, and more). In our work, we use a tool called \textit{mmore}~\cite{sallinen2025mmore}, a distributed pipeline similar to IBM's \textit{Docling}~\cite{docling}. The most useful feature of these tools is the ability to parse scanned documents, which we found was very valuable given that digitization in North Macedonia lags behind, and many available sources are scanned copies. Table \ref{tab:mk_sources} lists the sources processed using \textit{mmore}. All entities were contacted directly, and we obtained proper approval to use materials from each of them.

\begin{table}[!h]
\centering
\begin{tabular}{ll}
\hline
\textbf{Source} & \textbf{Origin} \\
\hline
Ss. Cyril and Methodius University in Skopje & \url{https://ukim.edu.mk/en/} \\
Macedonian Academy of Sciences and Arts & \url{https://manu.edu.mk/} \\
St. Clement of Ohrid University of Bitola & \url{https://uklo.edu.mk/?lang=en} \\
Goce Delčev University of Štip & \url{https://www.ugd.edu.mk/en/home/} \\
Institute of Macedonian Language & \url{http://imj.ukim.edu.mk/} \\
Official PE Gazette of North Macedonia & \url{https://www.slvesnik.com.mk/} \\
\hline
\end{tabular}
\caption{Macedonian Sources Processed with the Document-to-Text Pipeline}
\label{tab:mk_sources}
\end{table}

\subsection{Data for Instruction Model}
\label{app:instruct_data}
\begin{figure}[!h]
\centering
\begin{minipage}{0.45\textwidth}
    \centering
    \includegraphics[width=\linewidth]{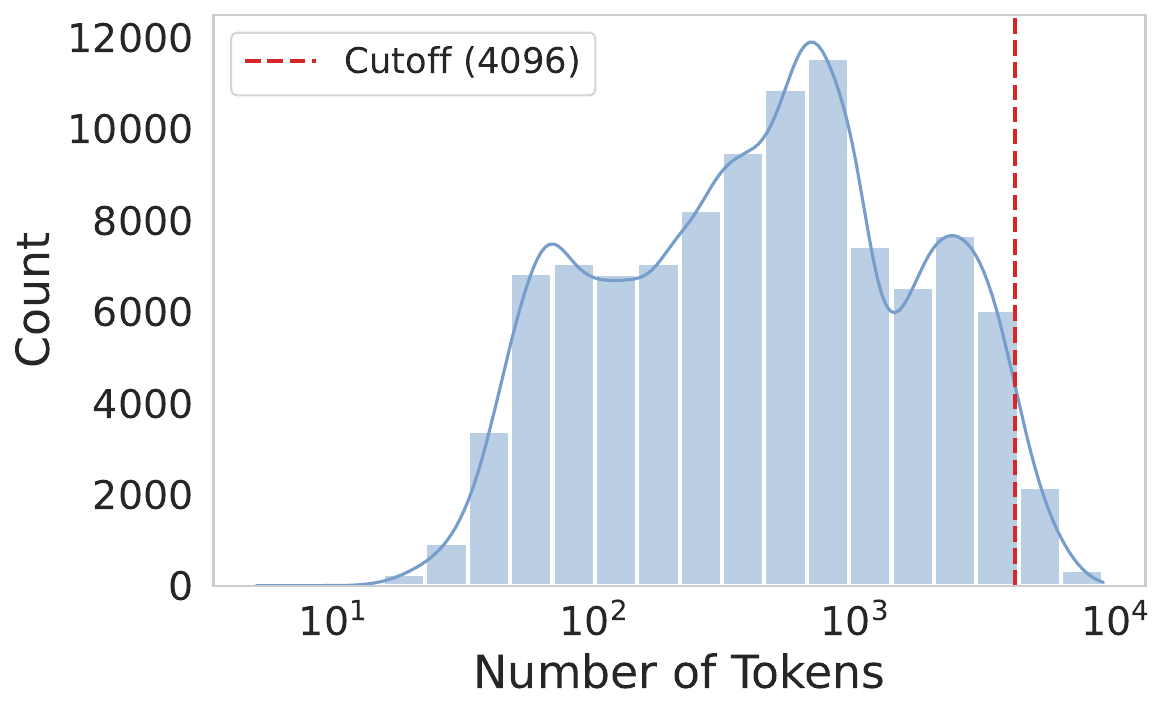}
    \caption{Token length distribution in the SFT dataset. The red dashed line indicates the 4,096-token cutoff, which covers 97.4\% of all samples.}
    \label{fig:cdf}
\end{minipage}
\hfill
\begin{minipage}{0.48\textwidth}
    \centering
    \begin{tikzpicture}
    \pie[
        text=legend,        
        radius=2.0,
        color={green1!50, purple1!50, orange1!50, yellow1!50},
    ]{
        58.5/Question Answering,
        33.0/Chat Conversations,
        5.3/Reasoning,
        3.2/Other (Essays+Code)
    }
    \end{tikzpicture}
    \caption{Distribution of Topics in the Instruction Dataset. Question Answering tasks comprise the majority (58.5\%), followed by Chat Conversations (33.0\%), with Reasoning and Other categories making up smaller portions (5.3\% and 3.2\% respectively).}
    \label{fig:instruction_topic_pie}
\end{minipage}
\end{figure}

Figure~\ref{fig:instruction_topic_pie} shows the composition of our instruction dataset across four high-level categories. The dataset is heavily dominated by \textit{question answering} and \textit{chat-style interactions}, which together account for over 80\% of all examples. A smaller portion is dedicated to \textit{reasoning tasks} and more open-ended formats such as \textit{code generation} and \textit{essay writing}, which help diversify the model's capabilities beyond straightforward instruction following.

Furthermore, Figure~\ref{fig:cdf} presents the token length distribution across the dataset. The majority of samples (97.4\%) fall below the 4,096-token cutoff used during supervised fine-tuning, ensuring that most examples are used without truncation.

\newpage
\subsection{System Prompt}
The system prompt that was used to train the instruction model is given below in both its original and English form.  

\begin{tcolorbox}[
  colback=green1!5,          
  colframe=green1!50!black,
  coltitle=white,
  fonttitle=\bfseries,
  title=System Prompt:,
  colbacktitle=green1!80!black,
  boxrule=0.4mm,          
  toptitle=1mm,
  bottomtitle=1mm
]

\textit{\textbf{Macedonian:} \foreignlanguage{russian}{Ти си виртуелен асистент кој помага на корисници на македонски јазик. Одговарај на прашања на јасен, разбирлив и професионален начин. Користи правилна граматика и обиди се одговорите да бидат што е можно покорисни и релевантни.
}}

\noindent\rule{14.8cm}{0.4pt}
\vspace{2mm}

\textit{\textbf{English:}} You are a virtual assistant that helps users in the Macedonian language. Answer questions in a clear, understandable, and professional manner. Use correct grammar and try to make your responses as helpful and relevant as possible.

\end{tcolorbox}

\subsection{Translation}
\label{sec:translation}

To preserve grammatical structure during translation of multiple-choice questions, we implement a \textit{template-based translation strategy}. Unlike naïve translation of isolated queries - which often produces grammatically flawed outputs - our approach maintains syntactic integrity through contextual grounding. Below we show the reason we went for this approach by using an example from the Serbian version of the ARC-Easy benchmark. 

\begin{tcolorbox}[
  colback=black!3,
  colframe=black!50,
  coltitle=white,
  fonttitle=\bfseries,
  title=\textbf{Example Instance from ARC-Easy:},
  colbacktitle=black!50,
  boxrule=0.4mm,
  toptitle=1mm,
  bottomtitle=1mm
]
\textbf{Original (Serbian):} Hladnokrvne životinje su često \\[1mm]
\textbf{Choices (Serbian):} ["brze", "velike", "bez dlake", "spore"] \\[1mm]
\textbf{Gloss (English):} Cold-blooded animals are often \\[1mm]
\textbf{Choices (English):} ["fast", "large", "hairless", "slow"]
\end{tcolorbox}

The naïve translation produces a grammatically awkward construction with syntactically incorrect word order, primarily due to missing subject-verb-object agreement.

\begin{tcolorbox}[
  colback=black!3,  
  colframe=black!50,      
  coltitle=white,    
  fonttitle=\bfseries, 
  title=\textbf{Naïve Translation (Incorrect)},
  colbacktitle=black!50,
  boxrule=0.4mm,
  toptitle=1mm,
  bottomtitle=1mm
]
\textbf{Translation (Macedonian):} \foreignlanguage{russian}{Често се ладнокрвни животни} \\[1mm]
\textbf{Gloss (English):} Often are cold-blooded animals
\end{tcolorbox}

To mitigate this issue, we insert a placeholder in place of the answer choice during translation, which is removed after processing. In addition to ensuring correct translation, this approach also helps prevent potential data leakage that could arise from choice-dependent translations.

\begin{tcolorbox}[
  colback=black!3,
  colframe=black!50,
  coltitle=white,
  fonttitle=\bfseries,
  title=\textbf{Template-Based Translation (Correct)},
  colbacktitle=black!50,
  boxrule=0.4mm,
  toptitle=1mm,
  bottomtitle=1mm
]

\textbf{Original (Serbian):} Hladnokrvne životinje su često \underline{\hspace{0.5cm}} \\[1mm]
\textbf{Translation (Macedonian):} \foreignlanguage{russian}{Ладнокрвните животни често се \underline{\hspace{0.5cm}}} \\[1mm]
\textbf{Gloss (English):} Cold-blooded animals are often \underline{\hspace{0.5cm}}

\end{tcolorbox}

This method ensures that sentence structure remains intact, avoiding artifacts introduced by out-of-context or partial sentence translations.   

\subsection{Survey Questions for Qualitative Evaluation}
\label{appendix:questions}

To assess the stylistic and cultural quality of model responses, we designed a set of ten original prompts reflecting everyday reasoning, communication, and local knowledge. Participants evaluated responses to the following questions (presented in English below):

\begin{tcolorbox}[
  colback=green1!5,
  colframe=green1!50!black,
  coltitle=white,
  fonttitle=\bfseries,
  title=Survey Questions:,
  colbacktitle=green1!80!black,
  boxrule=0.4mm,
  toptitle=1mm,
  bottomtitle=1mm
]

\begin{enumerate}[leftmargin=*, label=\arabic*.]
    \item What does it mean when someone says "It’s not going well for me"?
    \item How many years is compulsory education in North Macedonia?
    \item I have a headache. What medicine is recommended for relieving a headache?
    \item Tell me a good joke.
    \item Tell me a well-known Macedonian proverb.
    \item Write a short email postponing a meeting with a professor due to justified reasons.
    \item How would you respond to a friend who invites you for coffee, but you are busy?
    \item What should you do if you lose your ID card?
    \item What are the most famous tourist attractions in North Macedonia?
    \item If the bus is late and you have an important meeting, what is the smartest thing to do?
\end{enumerate}
\noindent\rule{14.8cm}{0.4pt}
\vspace{2mm}

\small{\textit{Participants were asked to evaluate responses based on fluency and relevance.}}

\end{tcolorbox}

\vspace{5mm}

To illustrate, we show example answers to Question 5 ("Tell me a well-known Macedonian proverb") generated by the two models:

\begin{tcolorbox}[
  colback=green1!5,
  colframe=green1!50!black,
  coltitle=white,
  fonttitle=\bfseries,
  title=Example Responses:,
  colbacktitle=green1!80!black,
  boxrule=0.4mm,
  toptitle=1mm,
  bottomtitle=1mm
]

\textbf{domestic-yak-instruct:}

\textit{\foreignlanguage{russian}{Една позната македонска изрека е: „Кој рано рани, две среќи граби“.}}

\vspace{2mm}
\textbf{Gloss (English):} A well-known Macedonian proverb is: "The one who rises early seizes two fortunes."

\vspace{1mm}
\noindent\rule{\linewidth}{0.4pt}
\vspace{1mm}

\textbf{Llama 70B Instruct:}

\textit{\foreignlanguage{russian}{Една поговорка која ми доаѓа на ум е: "Молчи и слушај, пајакот на ѕидот чувај." Оваа изрека значи дека човек треба да биде внимателен и да слуша, а не само да зборува. Слично на тоа, и другата изрека вели: "Ушите најпрво чујат, а потоа устата зборуваат." Овие изреки ни потсетуваат на важноста на слушањето и вниманието во нашиот секојден живот.}}

\vspace{2mm}

\textbf{Gloss (English):} One proverb that comes to mind is: "Be silent and listen, the spider on the wall beware." This proverb means that one should be attentive and listen, not just talk. Similarly, another proverb says: "The ears first hear, and then the mouth speaking." These proverbs remind us of the importance of listening and attention in our daily lives.

\end{tcolorbox}

\end{document}